\documentclass[10pt, conference, compsocconf]{IEEEtran}

\usepackage[pdftex]{graphicx}

\usepackage[cmex10]{amsmath}
\usepackage{amssymb}
\usepackage[font=footnotesize,caption=false]{subfig}
\usepackage{url}
\usepackage{etoolbox}
\apptocmd{\thebibliography}{\scriptsize}{}{}

\newcommand{\squeezeup}{\vspace{-2.5mm}}

\begin{document}

\title{Analysis of Convolutional Neural Networks for Document Image Classification}


\author{\IEEEauthorblockN{Chris Tensmeyer and Tony Martinez}
\IEEEauthorblockA{Dept. of Computer Science\\
Brigham Young University\\
Provo, USA\\
tensmeyer@byu.edu martinez@cs.byu.edu}
}

\maketitle

\begin{abstract}
Convolutional Neural Networks (CNNs) are state-of-the-art models for document image classification tasks.
However, many of these approaches rely on parameters and architectures designed for classifying natural images, which differ from document images.
We question whether this is appropriate and conduct a large empirical study to find what aspects of CNNs most affect performance on document images.
Among other results, we exceed the state-of-the-art on the RVL-CDIP dataset by using shear transform data augmentation and an architecture designed for a larger input image.
Additionally, we analyze the learned features and find evidence that CNNs trained on RVL-CDIP learn region-specific layout features.

\end{abstract}


\section{Introduction}

In recent years, Convolutional Neural Networks (CNNs) have been proposed for document image classification and have enjoyed great success.
These impressive results include: 
\begin{itemize}
\item 100\% accuracy on NIST tax forms dataset~\cite{nist91} using only one training sample per class~\cite{kang14}.
\item 89.8\% accuracy on the RVL-CDIP genre classification dataset.  The Bag of Words baseline was 49.3\%~\cite{harley15}.
\item 96.5\% accuracy in fine-grained classification of identity documents compared to 92.7\% accuracy using Histogram of Oriented Gradients (HOG) features~\cite{simon15}.
\end{itemize}

However, many such applications of CNNs to whole document image classification~\cite{harley15,afzal15,simon15} use a CNN pre-trained on the ImageNet dataset of natural images~\cite{russakovsky15} as a starting point for features.
Through this transfer learning approach is effective due to the generality of CNN features~\cite{sharif14}, there exist large domain differences between natural images and document images.
For example, in ImageNet, the object of interest can appear in any region of the image in a variety of 3D poses.
In contrast, many document images are 2D entities that occupy the whole image.
Due to such domain differences, we question whether the same architectures and techniques that are effective for natural images are also optimal for document images.

To answer this query, we conducted a large empirical study utilizing two large document image datasets.
To our knowledge, we are the first to conduct such a study for the domain of document images.
We examine many factors that contribute to CNN classification accuracy, which can be broadly categorized as data preprocessing and network architecture.
All CNNs in this work are randomly initialized and not pretrained on natural images.

For data preprocessing, we examine what representation(s) of the image (e.g. binary, RGB, HSV), are most effective as input to the network.
We also test 10 different types of label-preserving image transformations (e.g. crop, rotation, shear) to artificially expand the training data.
While cropping is typically applied in CNNs trained on ImageNet, we find that shear transforms yield best performance for document image tasks.
While not all document images are the same aspect ratio (AR), CNNs typically only accept inputs of a fixed size (e.g. 227x227) and hence a fixed AR.
We investigate this issue and find that CNNs trained with stochastic shearing are not adversely affected by AR warping of input images.

For network architectures, we examine factors such as network depth, width, and input size under various amounts of training data.
Critically, we achieve 90.8\% accuracy on RVL-CDIP \emph{without pretraining on natural images} by using a larger input size
This surpasses the previous best result of 89.8\%~\cite{harley15} on this dataset.
By incorporating multi-scale images into training and inference, we reach 91.03\%.
We also examine non-linear network operations in the domain of document images.

Lastly, we analyze what intermediate features CNNs learn from document image tasks.
Though CNNs are often considered black-box models, individual neurons can be characterized by their maximal-exciting inputs.
We find evidence that CNNs learn a wide variety of region-specific layout features.
Several intermediate neurons fire on page elements of specifics shapes and types (e.g. typed text, handwritten text, graphics).

\section{Related Work}

CNNs have been used in document image analysis for two decades for tasks such as character recognition~\cite{lecun98,simard03}, but only more recently were applied to large image classification tasks.
In 2012, Krizhevsky et al. showed the effectiveness of CNNs in large image classification by winning the ImageNet Large Scale Visual Recognition Challenge (ILSVRC) by a large margin~\cite{krizhevsky12}.
Since then, all top entries of ILSVRC have been based on CNNs, rather than on handcrafted features, with the 2015 winner surpassing human level performance by utilizing residual connections and batch normalization in a CNN with 152 layers~\cite{he15residual}.

In 2014, one of the first applications of CNNs to whole document image classification used a 4-layer CNN to classify tax forms and the Small Tobacco datasets~\cite{kang14}.
This CNN acheived 65.37\% accuracy on Small Tobacco compared to $\sim$42\% by the previous state-of-the-art HVP-RF classifier~\cite{kumar13}, which uses SURF local descriptors.

The following year, two works explored transferring the parameters of a CNN learned on ILSVRC to document classification tasks.
The DeepDocClassifier system~\cite{afzal15} retrained the top classification layer from scratch while finetuning the other layers.
They report an accuracy of 77.3\% on Small Tobacco.
Harley et al.~\cite{harley15} also transferred parameters from ILSVRC, but also introduced a larger dataset, called RVL-CDIP, that can be used to train CNNs from scratch.
They also found that a single holistic CNN outperformed an ensemble of region-specific CNNs on RVL-CDIP.

\section{Convolutional Neural Networks}

The CNNs we consider in this work are models that map input images $x \in \mathbb{R}^{H \times W \times D}$ into the probability vectors $y \in \mathbb{R}^C$, where $D$ is the input image depth (e.g. 3 for RGB) and $C$ is the number of classes.
Each layer, of the CNN performs an affine transformation with learnable parameters followed by non-linear operation(s):
\begin{equation} \label{eq:cnn}
x_{\ell} = g_{\ell}(W_{\ell} \star x_{\ell - 1} + b_{\ell})
\end{equation}
where $1 \leq \ell \leq L$ is the layer index, $x_0$ is the input image, $W_{\ell}, b_{\ell}$ are learnable parameters, $\star$ is either 2D convolution (with multi-channel kernels) for convolution layers or matrix multiplication for fully connected layers, and $g_{\ell}$ is a layer specific non-linearity, composed of $\operatorname{ReLU}(x) = \operatorname{max}(0, x)$, and optionally max-pooling, local response normalization, or dropout~\cite{krizhevsky12}.
The output of the last layer, $x_L$, is input to a softmax function, which outputs a probability vector over the target classes.
For more details, consult~\cite{krizhevsky12}.

Many of our experiments are based on the standard AlexNet architecture~\cite{krizhevsky12} (5 conv layers, 3 fully connected layers), but without the original sparse connections in the convolutional layers.
We also test many architectural variations, which are noted in the relevant experiments.

\section{Training Details}

\subsection{Datasets}

For this work, we utilize 2 datasets.
The first is the publicly avaialable RVL-CDIP dataset~\cite{harley15} which is composed of 400,000 grayscale images split into 320,000/40,000/40,000 train/val/test sets.
They are scanned office documents with 16 conceptual categories such as Letter, Memo, Email, Form.

The second dataset, denoted \emph{ANDOC}, is composed of genealogical records sampled from 974 collections owned by \url{Ancestry.com}. 
The target class of each image is its collection of origin (e.g. 1940 US Census, Pennsylvania death certificates 1880-1940).
In total, there are 880,000 images partitioned into a randomized 800,000/40,000/40,000 train/val/test split.
481,000 of these images are color, while the rest are grayscale.

For data preprocessing, pixel intensities are scaled to $[0,1]$ and then the channel mean is subtracted.

\subsection{Training hyperparameters}

We empirically determined good training schemes for each dataset.
For RVL-CDIP, CNNs are trained with Stochastic Gradient Descent (SGD) with mini-batches of 32 for 500,000 weight updates.
The initial learning rate (LR) is 0.003 and is decayed by a factor of 10 every 150,000 mini-batches.
For ANDOC, we used SGD with mini-batches of 128 for 250,000 weight updates.
The initial LR was 0.005 and is decayed by a factor of 10 every 100,000 mini-batches.
We believe the larger mini-batch does better for ANDOC because it has more output classes.

All networks are trained on the training split and progress is monitored on the validation set.  
The set of network parameters that performed best on the validation set is then evaluated on the test images.
In this work, we use test set accuracy as our evaluation metric.

CNNs require days to train on high-end GPUs, so training many CNNs for statistical significance testing is often too time-consuming.
For this reason, CNN literature almost always reports numbers from only a best single trained model.
Thus we typically report average performance of 1-2 CNNs.
However, in Section~\ref{sec:image_rep}, we trained 10 CNNs for each set of hyperparameters in order to measure the variance in model accuracy, which we estimate to be $\sigma \approx 0.1$ for RVL-CDIP and $\sigma \approx 0.05$ for ANDOC.

\section{Preprocessing Experiments}

\subsection{Image Representation}
\label{sec:image_rep}

\begin{figure}
\includegraphics[width=0.23\textwidth]{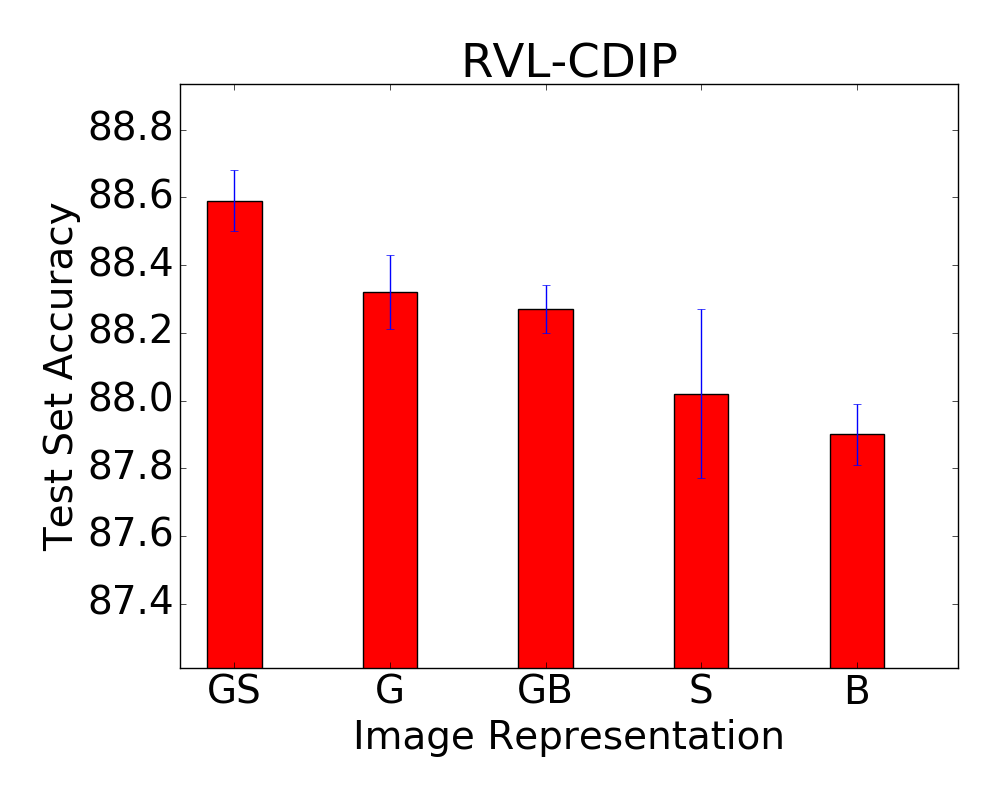}
\includegraphics[width=0.23\textwidth]{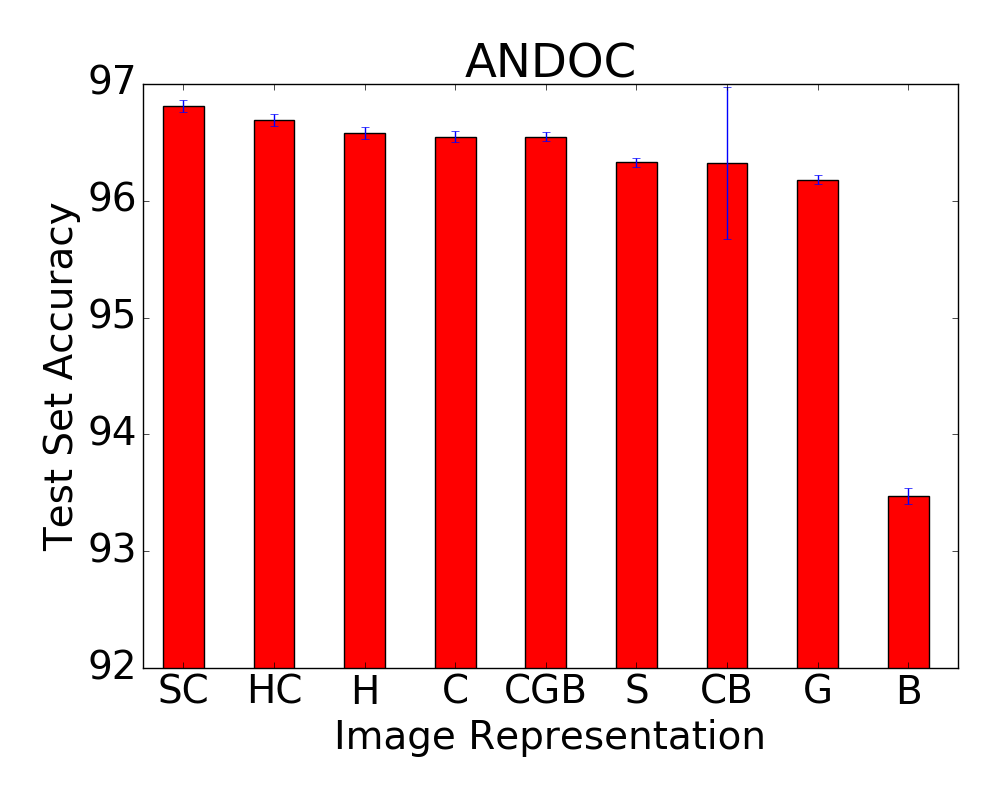}

\caption{Mean accuracy of 10 CNNs as we vary the image representation.  Error bars indicate the standard deviation of 10 trials. G=Grayscale, C=RGB, H=HSV, S=dense-SURF, B=Binary.}
\label{fig:image_rep}
\squeezeup
\end{figure}
	
Here we examine whether the best input representation is RGB, HSV, grayscale (G), binary (B), dense-SURF (S)~\cite{bay08}, or a combination.
For example, when combining RGB and G, we treat the input as a 4-channel image where channels 1-3 are RGB and channel 4 is G.
For D-SURF inputs, we compute SURF descriptors at fixed orientation on a 227x227 grid on the \emph{originally sized} images.
This results in a 64-dimensional descriptor for each grid point, which we treat as an image with 64 channels.
Binary images are computed with Otsu's method~\cite{otsu75}.

Figure~\ref{fig:image_rep} shows average accuracy of 10 CNNs.
Overall, combining RGB or G with S leads to the best accuracy.
Though S by itself does not perform well, we believe it combines well with RGB/G because the SURF descriptors are computed using the originally sized images.
As we later show, using larger input images lead to significant gains in performance.
Using B leads to worst performance and augmenting RGB/G with B leads to lower performance.
In ANDOC, the HSV colorspace performed equally as well as RGB, though combining both leads to a marginal increase.

For simplicity, the rest of the experiments use G input for RVL-CDIP and RGB input for ANDOC.
Additional experiments (results not shown) suggest that with optimal data augmentation (Section~\ref{sec:data_aug}), augmenting RGB/G with S does not lead to significant gains in performance.

\subsection{Data Augmentation}
\label{sec:data_aug}

\begin{table}

\begin{tabular}{l|c|c|c|c}

Transform                           & RVL-CDIP & RVL-CDIP & ANDOC   & ANDOC    \\
                                    & 1x Test  & 10x Test & 1x Test & 10x Test \\
\hline
None                                & 88.30    & 88.30    & 96.54   & 96.54    \\
Color Jitter                        & 88.36    & 88.34    & 96.43   & 96.41    \\
Crop                                & 88.83    & \textbf{89.31}    & 96.50   & 96.57    \\
Elastic~\cite{simard03}             & 87.94    & 88.12    & 96.27   & 96.36    \\
Gaussian Blur                       & 87.50    & 87.46    & 96.31   & 96.32    \\
Gaussian Noise                      & 88.19    & 88.18    & 96.57   & 96.47    \\
Mirror                              & 88.51    & 88.93    & 96.69   & \textbf{96.79}    \\
Perspective                         & 88.63    & 88.45    & 96.59   & 96.64    \\
Rotation                            & 88.74    & 77.46    & 96.37   & 96.36    \\
Salt/Pepper Noise                   & 88.03    & 87.97    & 96.39   & 96.45    \\
Shear                               & \textbf{89.33} & 84.88    & \textbf{96.75}   & 96.59    \\

\end{tabular}

\caption{Best performance for 10 types of data augmentation}
\squeezeup
\squeezeup
\label{tab:aug}
\end{table}

It is common practice to stochastically transform each input during SGD training to artificially enlarge the training set to improve performance~\cite{krizhevsky12,he15residual}.
We experimented with 10 types of transformations (e.g. crop, blur, rotation) and 2-4 parameter settings for each transformation type for a total of 38 CNNs trained with different data augmentation.
We report results for the best CNNs per transform type in Table~\ref{tab:aug}.
Following~\cite{simonyan14}, we also report multiple-view test performance where the overall CNN prediction is the average of the predictions made on 10 transforms of each test image.
This increases computation for prediction by 10x, but can increase accuracy~\cite{krizhevsky12}, which makes it appropriate for some applications.
The transforms used at test time are the same type as those used during training and include the untransformed image.
While some of these transforms are commonly used, shear, perspective, and elastic transforms have not been.

Shear transforms perform best for single-view testing and are comparable to the best multiple-view transforms.
When images are sheared (either horizontally or vertically), the relative locations of layout components are perturbed, but unlike rotation or perspective transforms, either the horizontal or vertical structure is preserved.
We believe that compared to other transform types, shearing best models the types of intra-class variations in document datasets.
The CNNs in the remaining experiments were trained using shearing with $\theta \in [-10^{\circ},10^{\circ}]$.

We then attempted to combine the two or three best performing types of transformations.
However, this did not improve performance over using a single transform for either single or multi-view testing (results omitted for brevity).

\subsection{Aspect Ratio}

\begin{figure}

\includegraphics[width=0.225\textwidth]{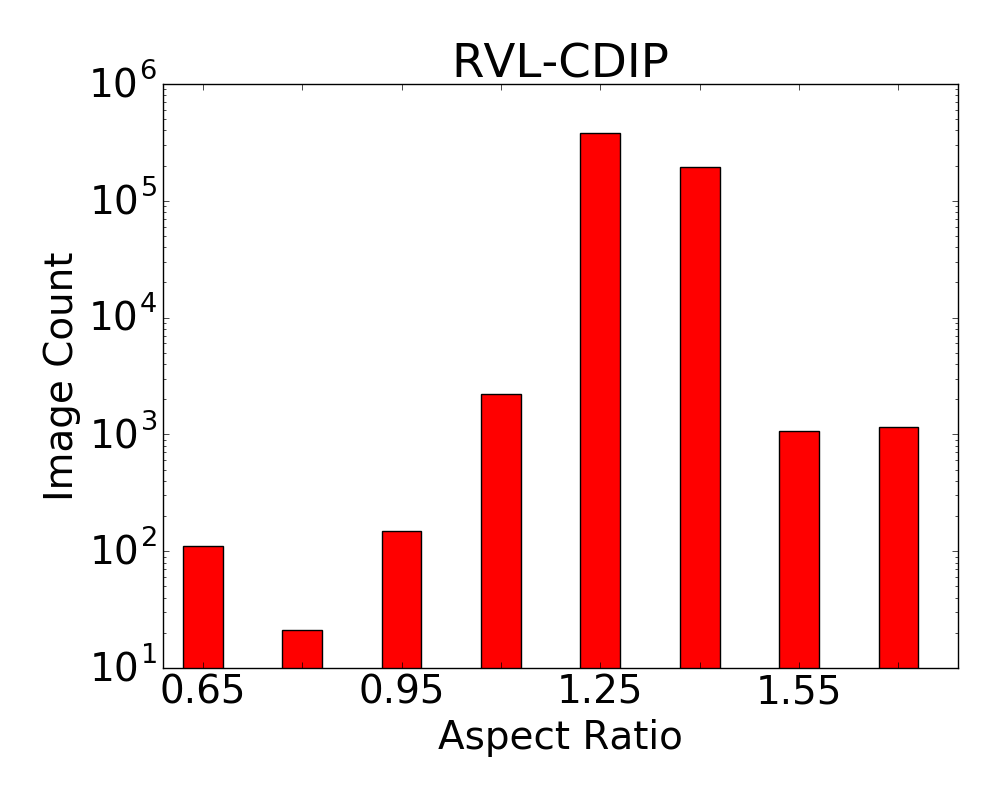}
\includegraphics[width=0.245\textwidth]{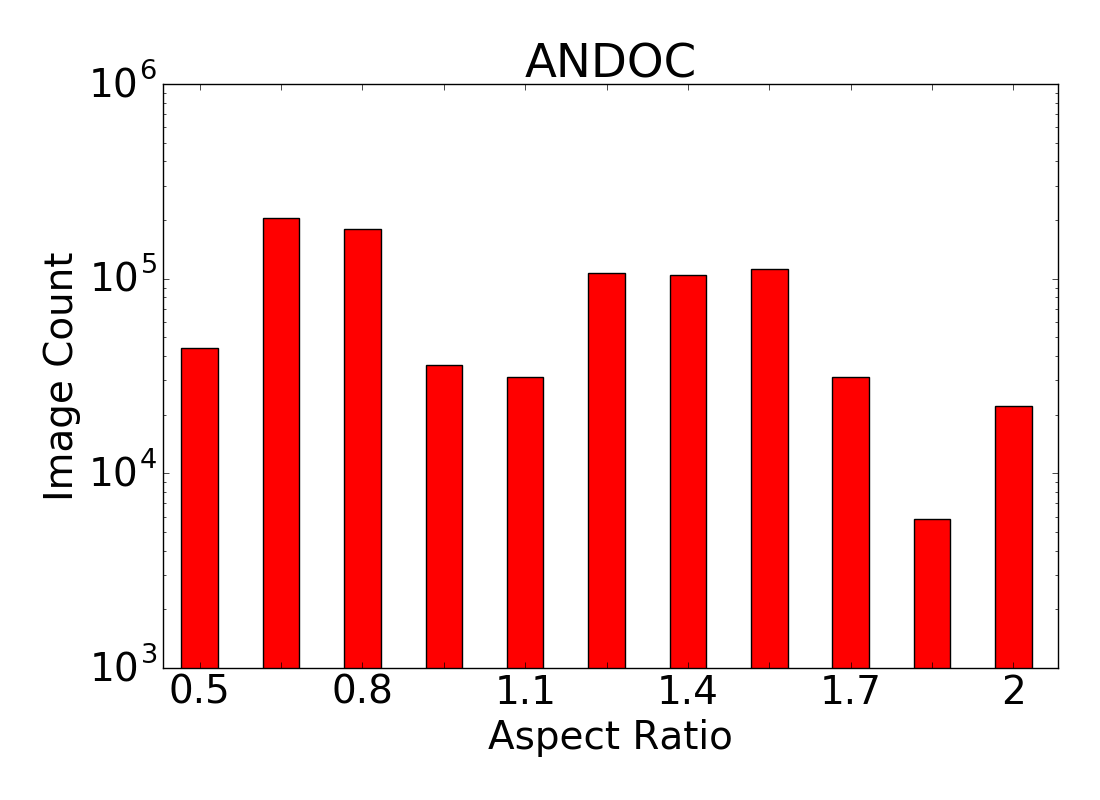}

\caption{Histogram of Aspect Ratios. Note y-axes are log scale.}
\label{fig:ar}
\end{figure}

\begin{table}
\begin{tabular}{l|c|c|c|c}

Method            & RVL-CDIP & RVL-CDIP & ANDOC   & ANDOC    \\
                  & 227x227  & 384x384  & 227x227 & 384x384  \\
\hline
Warped            &  89.25   &  90.84   & 96.73   & 97.14    \\
Padded            &  89.02   &  90.89   & 96.59   & 97.03    \\
Padded + SPP      &  88.85   &  \textbf{90.94}   & 96.59   & 97.12   \\
Warped  + SPP     &  \textbf{89.31}   & 90.92   & \textbf{96.77}   & \textbf{97.21}    \\
SPP (variable AR) &  88.51   &  88.71   & 96.20   & 97.15   \\
Variable AR + Crop$\dagger$ &  88.68   &  89.94   & 96.57   & 97.05    \\
\end{tabular}

$\dagger$ Predictions averaged over 3 crops.
\caption{Comparison of ways to address aspect ratio warping}
\label{tab:pad}
\squeezeup
\squeezeup
\end{table}

\label{sec:aspect_ratio}

One potential drawback of the standard CNN is fixed spatial input dimensions (e.g. 227x227).
This means all inputs must have the same aspect ratio (AR), though the original images may have various ARs.
For example, some document images are in portrait or landscape orientation or are printed on different paper sizes.
Figure~\ref{fig:ar} shows histograms of AR's in both datasets, which is more varied for ANDOC.
These images of various AR must be transformed to match the same expected CNN input dimensions.
A common technique resizes the image to the input size without maintaining AR.
We refer to this as \emph{inconsistent AR warping}.

In this experiment, we compare some alternatives to inconsistent AR warping in the context of document images.
One method pads images to the correct AR before resizing, however this wastes some precious input content.
Another method resizes (preserving AR) the image until the smallest dimension is the correct size.
Next, crops of the correct size are taken as input images.
At test time, predictions are averaged over 3 crops so the CNN sees all parts of the image.

A third way is to modify the CNN architecture to accept variable AR inputs.
While the convolution layers can operate on any input size, the fully connected layers expect a fixed sized vector input.
One way around the input size requirement is to replace the fixed 2x2 pooling regions of the last convolution layer with a Spatial Pyramid Pooling (SPP) operation~\cite{he15spatial}.
This way the size of the pooling regions vary with the input size and always yield the same sized vector output.
Inspired by~\cite{kumar13}, we experimented with pooling regions arranged in horizontal and vertical partitions (HVP), but it did not outperform SPP pooling, likely due to the small input size (13x13) at this layer.

We experimented with two different input sizes, 227x227 and 384x384.
The 384x384 CNN has the same overall structure as AlexNet, but is wider and has a greater amount of downsampling.
The variable AR inputs for the SPP-CNN were resized to have the same or fewer number of pixels as the fixed sized inputs.
In Table~\ref{tab:pad} we report the average test accuracy of 2 CNNs on each type of input.
Overall, no alternative outperformed inconsistent AR warping even with the SPP which accepts inputs of various AR.
Our hypothesis is that stochastically shearing input images (Section~\ref{sec:data_aug}) makes the CNN more invariant to inconsistent AR warping.

\section{Network Architecture Experiments}

\subsection{Depth}

\begin{figure}

\includegraphics[width=0.235\textwidth]{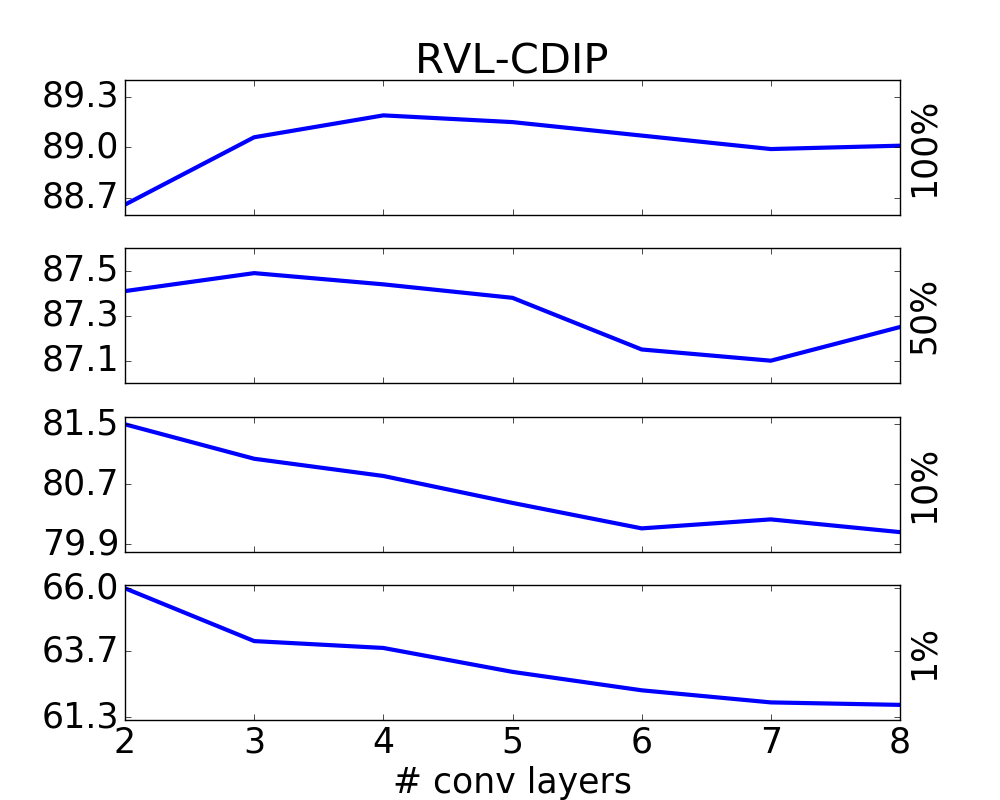}
\includegraphics[width=0.235\textwidth]{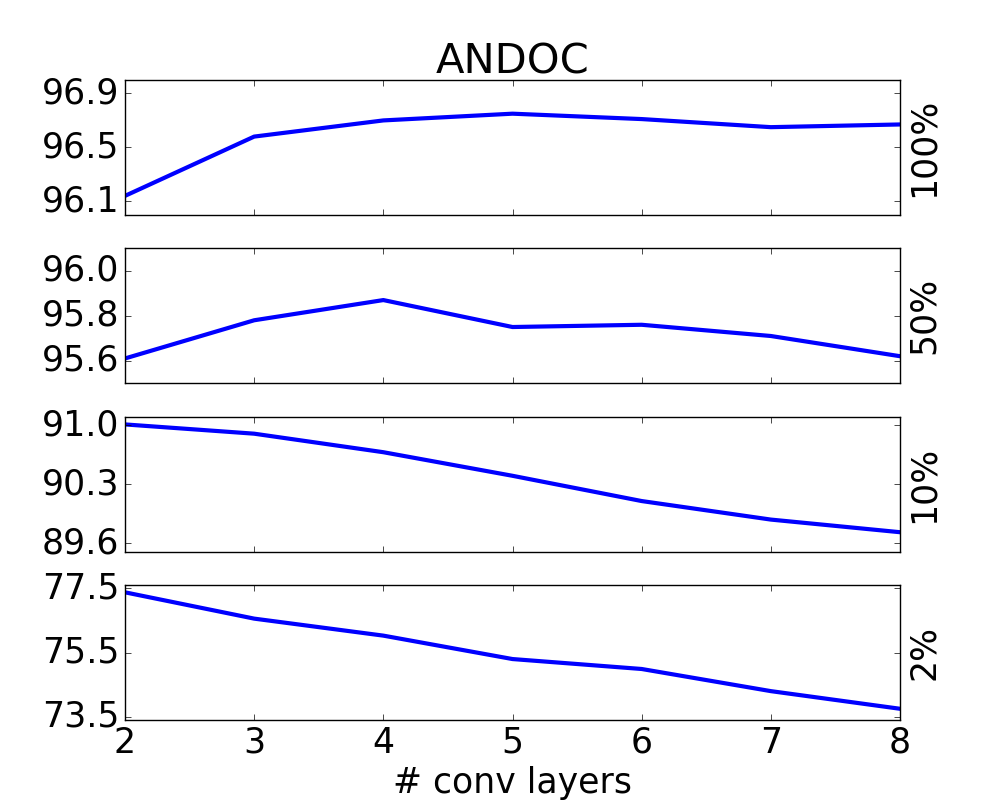}

\caption{Accuracy vs network depth.  Each sub-graph is for a different training set size, which is indicated to the right of the graph.}
\label{fig:depth}
\squeezeup
\end{figure}

The computational time of a CNN for both training and inference is influenced by the number of layers (depth) and the number of neurons in each layer (width).

To modify depth, we removed or added convolution layers to the AlexNet architecture, always keeping the first two layers and maintaining the same downsampling.
For shallower nets, we removed layers in this order: conv3, conv4, conv5.
For deeper nets, we inserted layers identical to conv3 between conv3 and conv4, where no pooling occurs.
We also ran these experiments where the size of the training set was reduced to 50\%, 10\%, 1\% for RVL-CDIP and 50\%, 10\%, 2\% for ANDOC.

Figure~\ref{fig:depth} gives the results for varying network depth.
Here we observe an overall decay in performance as depth increases, even for large amounts of training data.
This decay especially pronounced when less less than 10\% of the data is used ($<$32K or $<$79K instances), as increasing depth past 2 layers leads to sharp decreases in accuracy.
This validates the architectural choice of a 2-conv layer network of~\cite{kang14}.
These results suggest that network depth should adapt to the size of the training dataset.

We also observe the diminishing returns of additional data for CNN approaches, with only a 1-2\% decrease in performance with 50\% of the training data.
This suggests that gathering more data (e.g. millions of training examples) will only marginally improve CNN performance and more data efficient methods should be sought for.

\subsection{Width}

\begin{table}

\centering
\begin{tabular}{c|c|c|c|c}
Width & 100\% Data & 50\% Data & 10\% Data & 1\% Data  \\
\hline
10\%     & 82.80          & 82.70          & 77.95          & 62.12 \\
25\%     & 87.38          & 86.23          & 80.53          & \textbf{63.71} \\
50\%     & 88.75          & 87.26          & \textbf{80.68} & 63.16 \\
75\%     & 88.99          & \textbf{87.39} & 80.59 & 63.42 \\
100\%    & 89.18 & 87.15          & 80.52          & 63.48 \\
125\%    & 89.22 & 87.16          & 80.31          & 62.79 \\
150\%    & 89.20 & 87.22          & 80.46          & 62.98 \\
200\%    & \textbf{89.23} & 87.09          & 80.06          & 62.37 \\
\hline
fc-50\%  & \textbf{89.17} & \textbf{87.20} & \textbf{80.79} & \textbf{63.68} \\
conv-50\%& 88.78          & 87.09          & 80.61          & 63.24 \\
\end{tabular}

\caption{Accuracy of various width CNNs on RVL-CDIP with different amounts of training data}
\label{tab:width}
\squeezeup
\squeezeup
\end{table}

To modify width, we multiply the number of neurons in each layer by a constant factor.
We also experiment with changing just the width of the convolution layers or just the width of the fully connected layers.
Results for RVL-CDIP (ANDOC results were similar) are presented in Table~\ref{tab:width}.

With the full training set, network performance saturates at 100\% of the AlexNet width.
For smaller datasets, smaller widths are optimal.
We also see that reducing the width of only the convolutional layers leads to lower performance than just reducing the width of the fully connected layers.

\subsection{Input Size}

\begin{figure}

\includegraphics[width=0.235\textwidth]{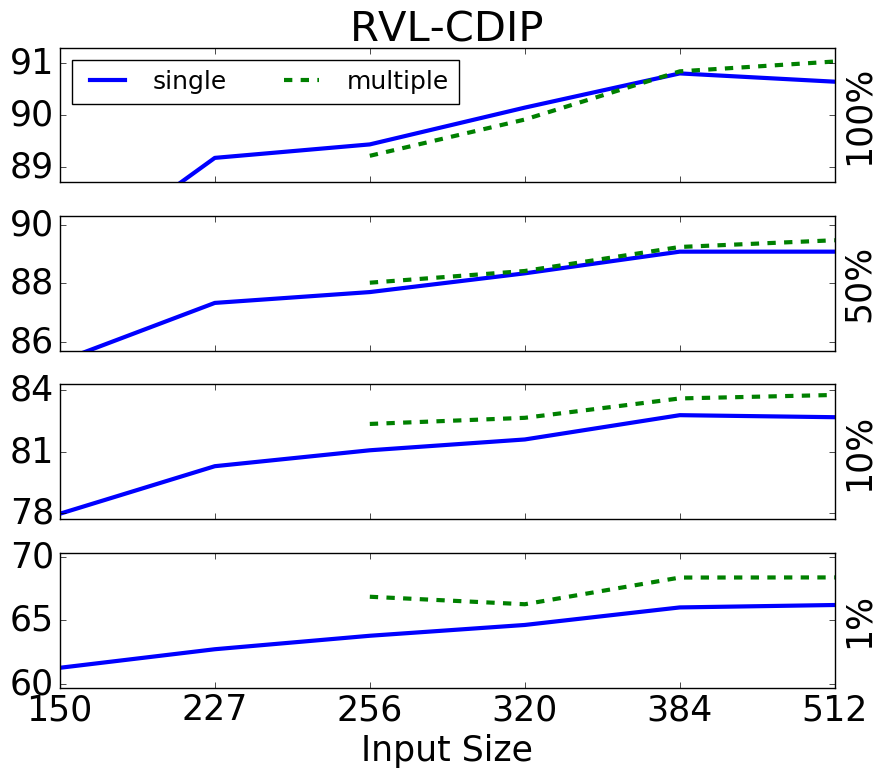}
\includegraphics[width=0.235\textwidth]{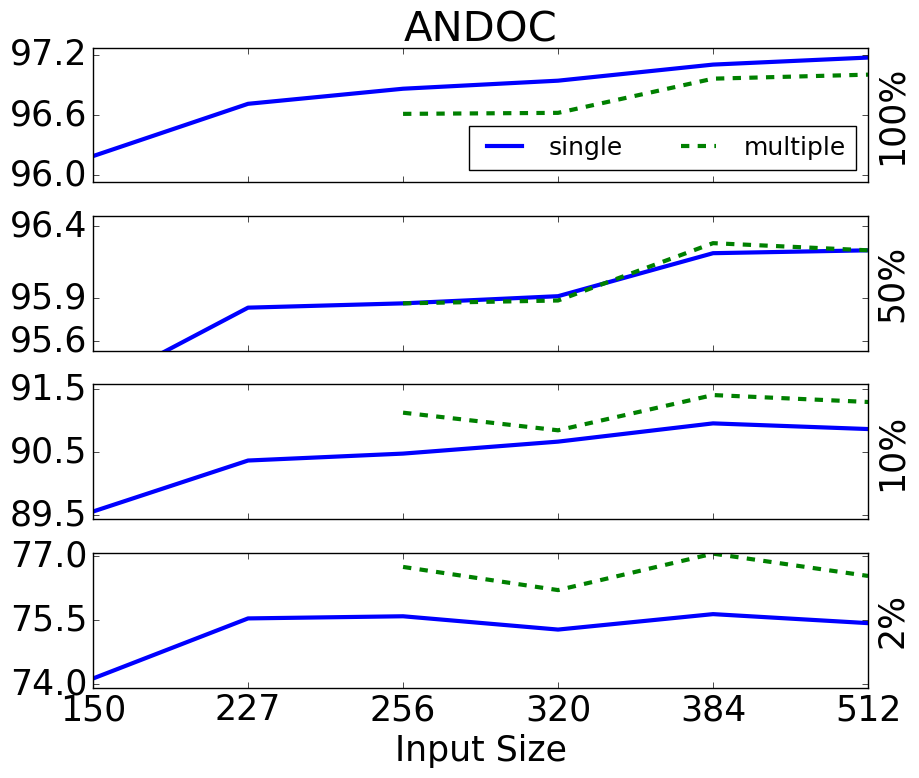}

\caption{Accuracy vs input size.  Solid blue lines are for CNNs trained with a single input size. Dashed green lines are for CNNs trained with variable sized input images.  The x-axis location corresponds to the largest input size used to train the CNN.  For example, the performance of the CNNs trained with images of size 256-384 is plotted at x=384.}
\label{fig:input_size}
\end{figure}

Standard CNN architectures~\cite{krizhevsky12,simonyan14,he15residual} accept inputs that are 224x224, but to our knowledge, this design choice has not been thoroughly explored before.
At this resolution, large text is generally legible, but smaller text is not.
We empirically test square input sizes $\{n\times{}n | n = 32,64,100,150,227,256,320,384,512\}$.
Of necessity, we modify the architecture of the CNN for each input size with the following principles.
\begin{enumerate}
\item Same number of layers
\item Increase kernel size and network width with input size
\item Spatial output size of final convolution layer is 6x6
\end{enumerate}
An exhaustive grid search over possible architectures was not possible with our computational resources and we acknowledge that our chosen architectures may not be optimal.
Figure~\ref{fig:input_size} shows a distinct trend with larger inputs leading to increased performance.
In fact, increasing the input size to 384x384 for RVL-CDIP yields an accuracy of 90.8\%, which exceeds the previous published best result of 89.9\%~\cite{harley15}.


Given that input size significantly impacts performance, we went further to evaluate multi-scale training and testing of CNNs on document images.
We did this by using the architecture of the largest input size and replacing the last fixed-sized pooling regions of the CNN with SPP pooling regions~\cite{he15spatial} (see Section~\ref{sec:aspect_ratio}).
During training, all images are randomly resized within a prespecified range (spanning 3 sizes in Figure~\ref{fig:input_size}).
At test time, we average predictions across 3 sizes and report the results in Figure~\ref{fig:input_size}.
This may be considered another type of data augmentation (Section~\ref{sec:data_aug}), but only works for architectures that can process multiple sizes.
This further improves performance on RVL-CDIP to 91.03\% for a CNN trained on images of size 320-512 and predictions averaged across sizes 320,384,512.
Smaller datasets also have higher relative improvement using multi-scale training and testing.

\subsection{Non-linearities}

\begin{table}
\centering
\begin{tabular}{l|c|c}

Non-linearity           & RVL-CDIP          & ANDOC    \\
\hline
LRN + Dropout + BN      &  \textbf{89.25}   & 97.45    \\
LRN + Dropout           &  89.21   & 96.72    \\
LRN + BN                &  89.03            & \textbf{97.57}    \\
LRN                     &  87.63            & 95.98    \\
Dropout                 &  89.17  & 96.72    \\
\end{tabular}

\caption{Accuracy based on Non-linearities}
\label{tab:nonlinear}
\squeezeup
\squeezeup
\squeezeup
\end{table}

Here we test whether LRN or Dropout non-linear components contribute to the overall CNN performance.
We also examine a relative new technique termed Batch Normalization (BN) that has been shown to both improve performance and increase training speed~\cite{ioffe15}.
BN works by first linearly scaling and shifting each neuron's activations to have zero mean and unit variance.
These statistics are calculated from the activations caused by just the images in each mini-batch.
This is then followed by a learnable shifting and scaling of the activation values so that the mean and variance of the neuron's activation values is learned.
We insert BN after each convolution or matrix multiplication and before ReLU.

Results are shown in Table~\ref{tab:nonlinear}.
For ANDOC, BN improves performance by a large margin and can replace dropout.
This is likely due to the visual variety of the documents (due to diverse set of classes) in ANDOC.
In contrast, Dropout is better than BN in RVL-CDIP, likely due to the visual uniformity of office-style documents.
We also observe that LRN also does not increase performance for either dataset.

%

\section{Analyzing what is learned}

\begin{table}
\centering

\begin{tabular}{|l|c|c|c|c|c|}
\hline
Category          & conv1 & conv2 & conv3 & conv4 & conv5 \\
\hline
Edge              & 10    & 2     & 0     & 0     & 0     \\
Parallel lines    & 46    & 9     & 10    & 9     & 2     \\
Checker           & 5     & 1     & 2     & 2     & 0     \\
Stroke(s)         & 10    & 10    & 2     & 0     & 0     \\
Shape Pattern(s)  & 15    & 32    & 24    & 14    & 16    \\
Corner            & 5     & 25    & 13    & 5     & 3     \\
Small Text        & 3     & 4     & 8     & 23    & 42    \\
Large Text        & 0     & 16    & 20    & 23    & 12    \\
Column/Margin     & 0     & 0     & 11    & 5     & 9     \\
Table Cells       & 1     & 0     & 2     & 0     & 2     \\
Graphic           & 0     & 1     & 4     & 4     & 0     \\
Handwritten Text  & 0     & 0     & 0     & 0     & 3     \\
Scanner Noise     & 0     & 0     & 0     & 0     & 3     \\
Ambiguous        & 0     & 0     & 2     & 15    & 8     \\
\hline
\end{tabular}

\caption{Categorization of 100 neurons for each convolution layer.}
\label{tab:neurons}
\squeezeup
\end{table}

\begin{figure}
\centering
\includegraphics[width=0.45\textwidth]{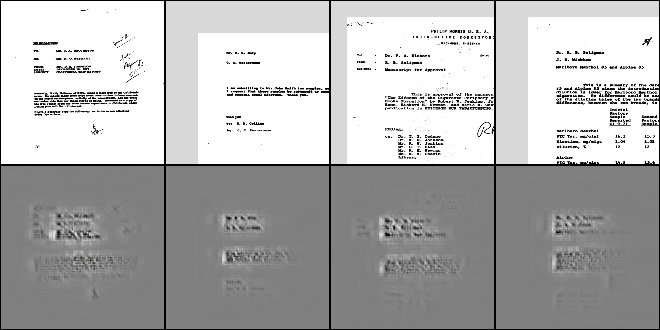}
\vspace{4pt}

\includegraphics[width=0.45\textwidth]{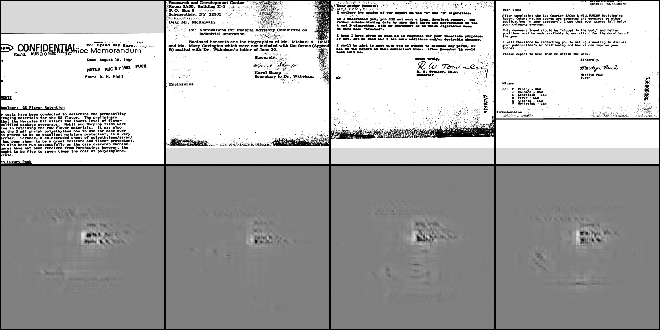}

\caption{Top 4 activating patches and deconv visualizations for 2 text detection neurons in conv5.  The top neuron detects left justified headers followed by wide paragraphs of text.  The bottom neuron detects right justified text below paragraphs of text.  Both features are class discriminative (e.g. \emph{Letter} class).}
\label{fig:text_detectors}
\squeezeup
\squeezeup
\end{figure}

\begin{figure}

\centering
\subfloat[conv1]{\includegraphics[width=0.15\textwidth]{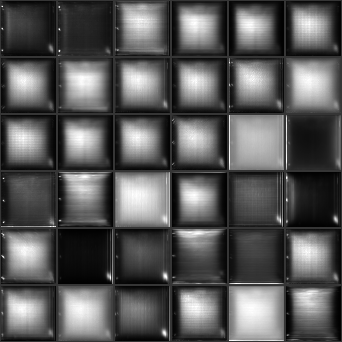}} \hspace{5pt}
\subfloat[conv2]{\includegraphics[width=0.15\textwidth]{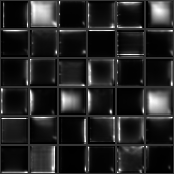}} \hspace{5pt}
\subfloat[conv3]{\includegraphics[width=0.15\textwidth]{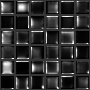}}

\subfloat[conv4]{\includegraphics[width=0.15\textwidth]{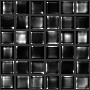}} \hspace{5pt}
\subfloat[conv5]{\includegraphics[width=0.15\textwidth]{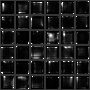}}

\caption{Average of filter responses for 36 filters for convolution layers.  Layers conv2-5 have filters that are specific to spatial regions.}
\label{fig:spatial}
\squeezeup
\end{figure}

One critique of CNNs is that they lack interpretation, making it difficult to know what features they use for classification decisions.
We analyzed a CNN trained on RVL-CDIP by examining the top-9 input patches that excite neurons and the \emph{deconv} visualization of each of these patches~\cite{zeiler14}.
The deconv visualization runs the network backwards from the neuron to the input in the context of some input image or patch.
It highlights the salient parts of the image that led to the neuron firing.
We found this technique critical for deciphering patterns recognized by neurons in later layers because the effective input size (i.e. receptive field) is very large.
For this analysis, we utilized the visualization tool proposed in~\cite{yosinski15}.

In our analysis we manually categorized the first 100 neurons in each convolutional layer (see Table~\ref{tab:neurons}).
Although the category of some neurons is subjective,
our labeling captures the trends in the features learned by each layer.
As expected, the first two layers detect simple elements which are gradually abstracted into complex detectors for text and arrangement of elements.
While our category labels were the same for all layers, the complexity of each category increased for deeper layers.
It is noteworthy that almost all neurons were interpretable, with the majority of the \emph{Ambiguous} category being neurons that fired on two distinct types of features.
This was not the case for the 4-layer network of~\cite{kang14}, whose first layer filters appear random.

Neurons in conv5 tended to find particular configurations of text, such as those shown in Figure~\ref{fig:text_detectors}.
One interpretation is that the CNN is performing a loose form of layout analysis as an intermediary step to classification.
The fully connected layers can then reason about the spatial correlation of these elements to form a classification decision.
We also observed that by conv5, the CNN has learned to distinguish between handwritten text and type-set text, though no explicit information about the type of text was provided to the CNN.

We also examined the spatial specificity of features by averaging the intermediate output images for each filter across all images of RVL-CDIP.
As seen in Figure~\ref{fig:spatial}, conv1 filter responses are generally not confined to any portion of the image.
However, the other layers exhibit many filters that only respond to certain regions, which is consistent with the hypothesis that CNNs learn layout features.

\section{Conclusion}

We examined several factors that influence CNN performance on document images.
Overall, applying shear transforms during training and using large input images lead to the biggest gains in performance, acheiving state-of-the-art performance on RVL-CDIP at 90.8\% accuracy.
Multi-scale training and testing also improve performance, specifically for smaller training sets.
As well, BN is a useful alternative to Dropout in datasets that have large visual variety.

We also examined a CNN trained on RVL-CDIP and found evidence that the CNN is learning intermediate layout features.
Neurons fire based on type of layout component (graphic, text, handwriting, noise, etc) and tend to fire on specific locations on the image.

{\tiny
\newcommand{\BIBdecl}{\setlength{\itemsep}{0.25 em}}
\bibliographystyle{IEEEtran}
\bibliography{bib}

}
	
\end{document}